\UseRawInputEncoding

\documentclass{article}

\usepackage{microtype}
\usepackage{graphicx}
\usepackage{subfigure}
\usepackage{booktabs} 
\usepackage{color}
\usepackage{amsfonts}
\usepackage{amssymb} 
\usepackage{bm}
\usepackage{amsmath,amsthm,amsfonts,amssymb,amscd}
\usepackage{csquotes}
\usepackage{bbm}
\usepackage{xcolor}

\usepackage{hyperref}

\usepackage[accepted]{icml2021}

\icmltitlerunning{Evaluating Saliency Map Explanation on Multi-Modal Medical Images}

\begin{document}

\twocolumn[
\icmltitle{One Map Does Not Fit All: \\
Evaluating Saliency Map Explanation on Multi-Modal Medical Images}



\begin{icmlauthorlist}
\icmlauthor{Weina Jin}{to}
\icmlauthor{Xiaoxiao Li}{goo}
\icmlauthor{Ghassan Hamarneh}{to}
\end{icmlauthorlist}

\icmlaffiliation{to}{School of Computing Science, Simon Fraser University, Burnaby, Canada}
\icmlaffiliation{goo}{Department of Electrical and Computer Engineering, The University of British Columbia, Vancouver, Canada}

\icmlcorrespondingauthor{Weina Jin}{weinaj@sfu.ca}

\icmlkeywords{Interpretable Machine Learning, Medical Imaging Analysis}

\vskip 0.3in
]



\printAffiliationsAndNotice{}  

\begin{abstract}
Being able to explain the prediction to clinical end-users is a necessity to leverage the power of AI models for clinical decision support. For medical images, saliency maps are the most common form of explanation. The maps highlight important features for AI model's prediction. Although many saliency map methods have been proposed, it is unknown how well they perform on explaining decisions on multi-modal medical images, where each modality/channel carries distinct clinical meanings of the same underlying biomedical phenomenon.
Understanding such \textit{modality-dependent features} is essential for clinical users' interpretation of AI decisions.
To tackle this clinically important but technically ignored problem, we propose the MSFI (Modality-Specific Feature Importance) metric to examine whether saliency maps can highlight modality-specific important features. MSFI encodes the clinical requirements on modality prioritization and modality-specific feature localization.
Our evaluations on 16 commonly used saliency map methods, including a clinician user study, show that although most saliency map methods captured modality importance information in general, most of them failed to highlight modality-specific important features consistently and precisely. The evaluation results guide the choices of saliency map methods and provide insights to propose new ones targeting clinical applications. 
\end{abstract}

\begin{figure}[h]
    \centering
    \includegraphics[width=0.9\linewidth]{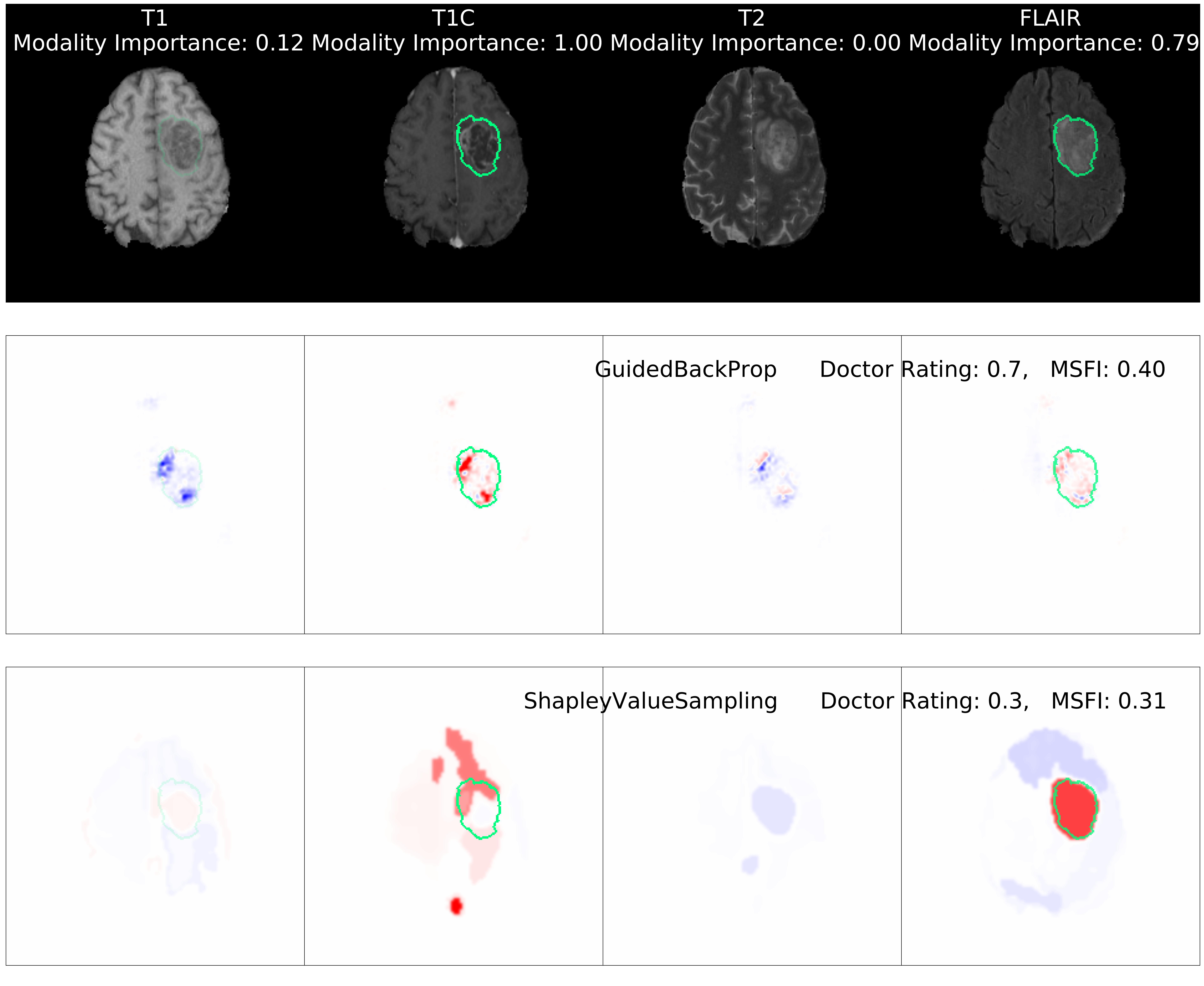}
    \caption{\textbf{Multi-modality MRI and its saliency maps.} The MRI shows a high-grade glioma. Although the brain tumor is visible on all modalities/pulse sequences of T1, T1C, T2 and FLAIR, multiple modalities present different signals which portray various compositions in the tissue. The radiographic features of contrast enhancement is most visible on T1C modality as hyperintense signal, and vasogenic edema appears as hyperintense signal abnormality on T2/FLAIR~\cite{uptodate}. Green contour is the ground-truth tumor segmentation mask, with its intensity indicating the degree of importance a modality (Modality Importance) to model's prediction. We show saliency maps of a gradient-based method (GuidedBackProp) and a perturbation-based method (ShapleyValueSampling), with red color indicating the most important features, and blue the least important. We also present our proposed saliency map evaluation score --- MSFI score and clinician rating from our user study, both rated on 3D images with $[0,1]$ range.}
    \label{fig:hm}
\end{figure}

\begin{figure*}[!tb]
    \centering
    \includegraphics[width=0.99\textwidth]{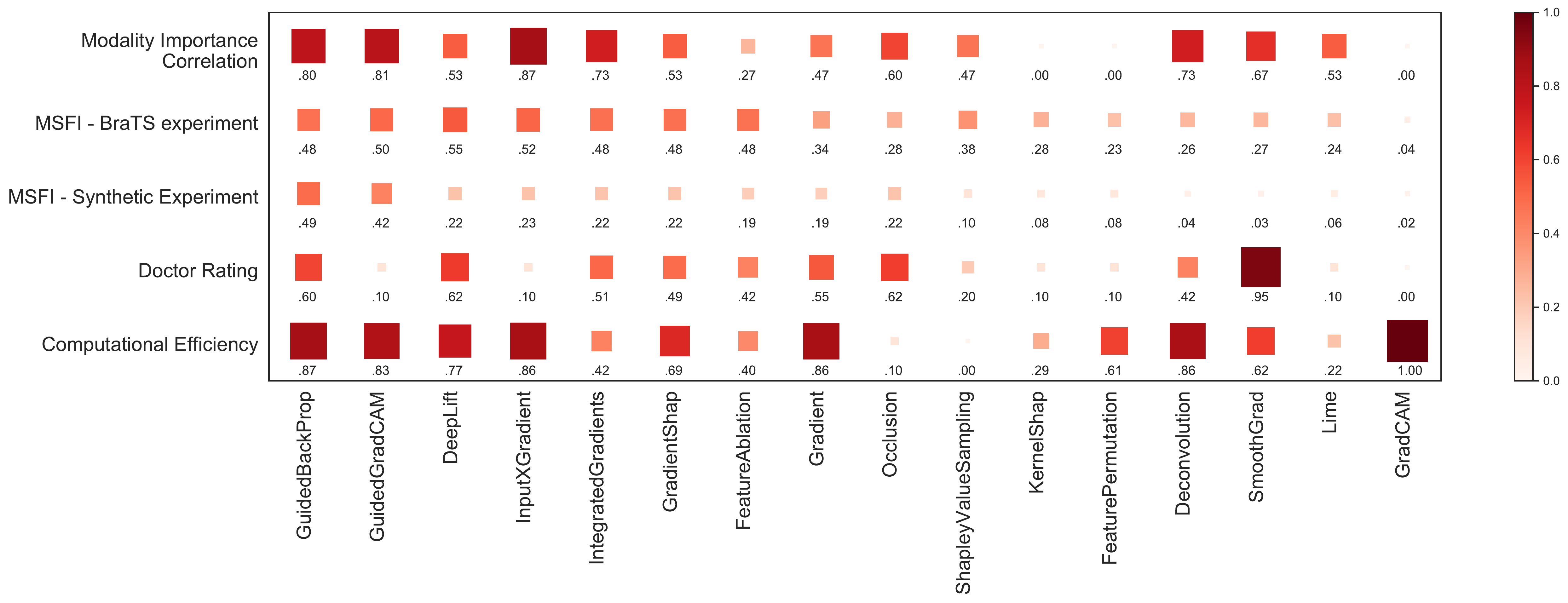}
    \caption{\textbf{Comparison matrix of the evaluated 16 saliency map methods.} The rows are evaluation metrics, and the columns are saliency map methods. All metrics are normalized to $[0,1]$, and is indicated by the size of the square or the color. The darker or bigger a square is, the better a method is in that metric. We sort the methods in descending order according to their two summed MSFI scores (Row $2 + 3$).}
    \label{fig:results}
\end{figure*}

\section{Introduction}\label{intro}
Being able to explain decisions to users is a sought-after quality of deep learning-based models, particularly when considering their application in clinical decision support systems~\cite{Jin_2020,He2019,Kelly2019}. Explanations can help users verify the artificial intelligence (AI) model's decision~\cite{Ribeiro2016b}, calibrate their trust in AI assistance~\cite{Zhang2020,7349687}, identify potential biases~\cite{Caruana2015},
 make biomedical discoveries~\cite{Woo2017}, meet ethical and legal requirements~\cite{Amann2020,gdpr}, and ultimately facilitate doctor-AI communication and collaboration to leverage the strengths of both~\cite{2101.01524,Topol2019,Carter2017}. 
 Prior user studies with physicians show that doctors tend to check the explanation to understand AI's results when results are relevant to their own hypothesis or differential diagnosis~\cite{Xie2020}, or to resolve conflicts when they disagree with AI~\cite{10.1145/3359206}.

Regarding medical imaging-related tasks, the most common and clinical end-user-friendly explanation is a saliency map~\cite{jin2021euca,Reyes2020,Singh2020}. It is also referred to, in the literature, as \textit{sensitivity map}, \textit{heatmap}, \textit{feature attribution map}, or \textit{feature importance map}. Saliency map methods are a group of feature attribution-based methods. They highlight image regions that are important for the model's prediction, and encode the important scores for each image pixel by a color map overlaid on the input image (Fig.~\ref{fig:hm}).
A recent user study involving clinicians showed that knowing the important features that drive AI's prediction is crucial, especially when clinicians need to compare AI's decision to their own clinical judgment in cases of decision discrepancy~\cite{pmlr-v106-tonekaboni19a}. 
Many methods have been proposed to generate saliency maps in the explainable AI (XAI) and computer vision communities. These methods were mainly designed for and evaluated on natural images. Given the discrepancies between natural and medical images, the question of how well the saliency map methods explain decisions made on
medical images, remains understudied.

One major discrepancy between natural and medical images is that a medical imaging study could include multiple image modalities or channels~\cite{MartBonmat2010} and such modality-specific information is clinically meaningful. 
For example, different pulse sequences of magnetic resonance imaging (MRI) technique — T1 weighted, T2 weighted, or fluid-attenuated inversion recovery (FLAIR) modalities;  dual-modality imaging of positron emission tomography-computed tomography (PET-CT)~\cite{pmid12072843}; CT images viewed at different levels and windows to observe different structures such as bones, lungs, and other soft tissues~\cite{HARRIS1993241}; 
multi-modal endoscopy imaging~\cite{Ray2017}; Photographic (also called clinical), dermoscopic, and hyper-spectral images of a skin lesion~\cite{8333693,Zherebtsov2019}; multiple stained microscopic or histopathological images~\cite{Long2020, Song2013}.
The signals from multiple image modalities capture different aspects of the same underlying cells, tissues, lesions, or organs.

Doctors rely on cross-modality comparison and combination to reason about a diagnosis or differential diagnosis. For instance, in a radiology report on MRI, radiologists usually observe and describe \textit{anatomical} structures in T1 modality, and \textit{pathological} changes in T2 modality~\cite{cochard_netter_2012, Bitar2006}; doctors can infer the composition of a lesion (such as fat, hemorrhage, protein, fluid) by combining its signals from different MRI modalities~\cite{Patel2016}. In addition, some imaging modalities are particularly crucial for the diagnosis and management of certain diseases, such as a contrast-enhanced modality of CT or MRI in a tumor case, and diffusion-weighted imaging (DWI) modality MRI to a stroke case~\cite{Lansberg2000}. 

\begin{table*}[!ht]
    \centering
    \begin{tabular}{p{0.18\linewidth} |p{0.22\linewidth}|p{0.23\linewidth}|p{0.22\linewidth}}
\toprule
\textbf{Method} & \textbf{Require access to model parameters} & \textbf{Model dependency} & \textbf{Computational efficiency}\\
\hline
Activation-based & Yes & Deep neural networks & Fast\\
\hline
Gradient-based & Yes & Differentiable models & Fast\\
\hline
Perturbation-based & No &Model-agnostic & Slow\\
\bottomrule
\end{tabular}
    \caption{\textbf{Comparison of three groups of post-hoc saliency map methods}}
    \label{tab:method_group}
\end{table*}

The uniqueness of multi-modal medical images brings up new and clinically relevant requirements for saliency map methods: 
The explanation needs to support the clinical user's reasoning process towards decisions made based on multi-modal medical images, as stated above, and (to a certain degree) align with expert prior knowledge on what constitutes an important feature for each modality.
Given the scarcity of work on saliency map evaluation for medical images, it is unclear whether existing methods will fulfill such requirement, and what needs to be improved in current methods to ensure their suitability for multi-modal medical images.
One of the most conspicuous and clinically important questions to ask is: 
\begin{center}
\textit{Can the existing saliency map methods capture modality-specific important features that are aligned with clinical prior knowledge?}
\end{center}


In this work, we give an exploratory answer to this question by proposing evaluation methods and conducting extensive comparisons with 16 commonly used saliency map methods. Specifically, we propose two evaluation methods/metrics, Modality Importance (MI) and Modality-Specific Feature Importance (MSFI), that reflect clinical knowledge on interpreting multi-modal medical images.
The first metric, MI, corresponds to clinicians' prioritization of multiple modalities. It uses Shapley value~\cite{RM-670-PR} to determine the ground-truth importance for each image modality, and compares modality-wise summations of saliency map values with such a ground-truth. 
The second metric, MSFI, measures fine-grained clinical requirement on localizing features that are specific to each modality. It combines MI with feature localization masks as the ground-truth, and calculates the ratio of saliency map regions inside the segmentation map, weighted by MI. With the two metrics and a clinician user study, we conducted extensive evaluations on 16 saliency map methods that cover the most common activation-, gradient-, and perturbation-based approaches. The findings show that most saliency map methods can reflect the importance of a modality as a whole, but only a few methods could capture the modality-specific fine-grained features that meet clinical needs (Fig.~\ref{fig:results}). 
Our contributions in this work are: 
\begin{enumerate}
    \item We propound the clinically important problem of modality-specific saliency map explanation to the technical community; 
    \item We propose an evaluation metric MSFI that specifies clinical requirements on saliency maps from our physician user study. It assesses a given saliency map method's ability to capture modality-specific features deemed important for an AI model's decision. MSFI can be applied to any model and dataset with ground-truth feature localization masks only on test data.
    \item We conduct extensive evaluation, including a doctor user study, on 16 commonly used saliency map methods. The assessments provide insights for XAI method selection and for proposing new XAI methods targeting clinical decision support systems that rely on multi-modal medical images. 
\end{enumerate}

\section{Background and Related Work}
\subsection{Multi-Modality Learning}
Approaches for building convolutional neural network (CNN) models that fuse multi-modal medical images can be divided into three categories: methods that fuse multi-modal features at the \textit{input}-, \textit{feature}-, and/or \textit{decision}-levels~\cite{10.1007/978-3-030-32962-4_18}. We focus on the most common setting for multi-modal medical imaging learning tasks: \textit{input}-level multi-model image fusion~\cite{Shen2017}, in which the multi-modal images are stacked as image channels and fed as input to a deep convolutional neural network. The modality-specific information is fused by summing up the weighted modality value in the first convolutional layer.

\subsection{Saliency Map Methods}
We focus on saliency map methods that are \textbf{\textit{post-hoc}}. This group of methods is a type of proxy models that probe the model parameters and/or input-output pairs of an already deployed or trained black-box model. In contrast, the \textit{ante-hoc} saliency map methods -- such as attention mechanism -- are predictive models with explanations baked into the training process.
We leave out the \textit{ante-hoc} methods because such explanations are entangled in its specialized model architecture, which would introduce confounders in the evaluation.



We describe the three major approaches to generating post-hoc saliency maps and how they have been applied to medical imaging tasks. We review individual methods in each category in the Appendix. The key comparisons are highlighted in Table~\ref{tab:method_group}.



\textbf{Activation-based methods}, such as CAM (Class Activation Mapping)~\cite{7780688}, were proposed particularly for deep learning-based models. They aggregate neuron activation signals in a deep neural network, and visualize the signals in input space. Activation-based methods have been widely applied in medical imaging classification tasks such as chest X-Rays~\cite{Wang2017a,Rajpurkar2018,Zech2018} and knee MRIs~\cite{Bien2018} to reveal what the models had learned and potential biases.

\textbf{Gradient-based methods} calculate the partial derivative of a prediction with respect to an input feature. 
The application of this group of methods to medical imaging tasks is also quite popular, such as for physicians' quality assurance in brain tumor grading task~\cite{Meier2018}, for image biomarker discovery in neurodevelopment~\cite{KAWAHARA20171038}, for feature visualization on echocardiograms view classification~\cite{Madani2018fast}.

\textbf{Perturbation-based methods} calculate the marginal probability of a feature as its attribute score by sampling from perturbed inputs. Since this class of methods need only probe the input-output pairs to generate an explanation, they do not require access to model parameters and thus are applicable to any black-box models. But they are usually computationally intensive as they require multiple passes through the model to acquire the perturbed input-output samples. Existing works applied different perturbation sampling methods such as occlusion on optical coherence tomography (OCT) images~\cite{Kermany2018}, and Shapley values for Autism Spectrum Disorders biomarker discovery on MRI~\cite{Li2019a}.

In prior works, the saliency maps were usually presented as the sanity check of a model's performance. Also, most of the saliency maps were on single-modal medical imaging-related tasks. Research work is lacking when the goal is to study a saliency map's ability to capture modality-specific features, or to perform empirical evaluation of saliency maps for medical images.

\begin{figure*}[!tb]
    \centering
    \includegraphics[width=0.9\textwidth]{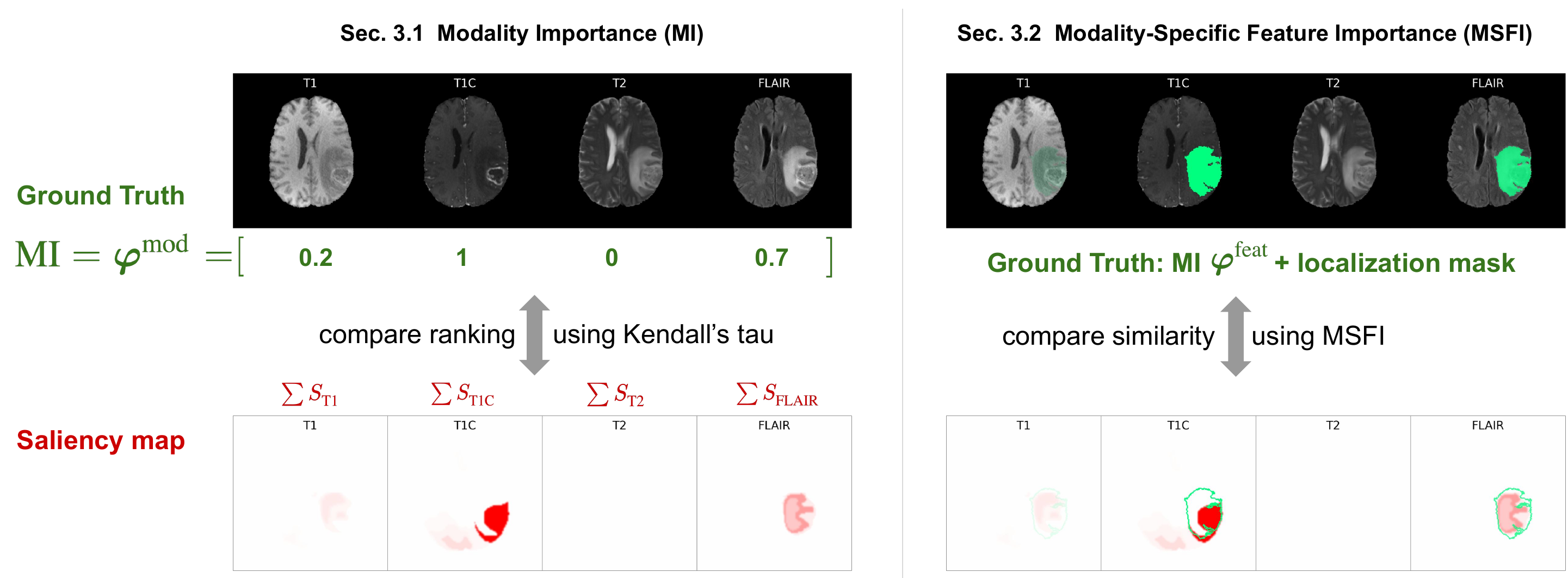}
    \caption{\textbf{The computational evaluation methods.}}
    \label{fig:eval_outline}
\end{figure*}

\subsection{Saliency Maps Evaluation}
Evaluating explanation methods is challenging, because the ground-truth explanation is human users' judgment. Such ground truth explanations are subjective and expensive to acquire especially from domain experts. To balance this, Doshi-Velez and Kim proposed three levels of evaluation: from the least expensive computational evaluation on proxy tasks, to human evaluation involving lay persons on simple tasks, to the most expensive domain expert evaluation on real tasks~\cite{doshivelez2017rigorous}. Our evaluations cover both computational and domain expert evaluation. 

Prior work proposed several computational evaluation metrics: Adebayo et al. proposed a sanity check on gradient-based explanations to evaluate their sensitivity to the randomization of training labels or model parameters~\cite{NEURIPS2018_294a8ed2}. Among the 7 methods they inspected, only \textit{Gradient} and \textit{SmoothGrad} pass the sanity check. 
Other evaluation metrics such as \textit{infidelity}~\cite{NEURIPS2019_a7471fdc} and \textit{Remove And Retrain}~\cite{DBLP:conf/nips/HookerEKK19} measure the changes in the output when perturbing the input. 
Object segmentation masks or bounding boxes can be used as a substitute ground-truth to quantify the localization capability of the saliency maps. Metrics such as intersection over union (IoU) are used to compare such a ground truth with the calculated saliency map~\cite{10.1007/978-3-030-32226-7_82,netdissect2017}.
Kim et al. conducted a controlled experiment on synthesized images with different levels of agreement between the ground-truth label and spurious features~\cite{pmlr-v80-kim18d}; their work inspired our synthetic experiment (Section~\ref{syn}).


\section{Methods to Evaluate Multi-Modal Saliency Map Explanations}

The overarching goal of the evaluation is to examine whether a saliency map method can reliably reveal features on multi-modal medical images that are clinically meaningful, given a well-trained model. Based on our physician user study (Section~\ref{user_study}), we abstract such goal as two specific clinical requirements: modality prioritization, and modality-specific feature localization, which correspond to our evaluations at two granularity levels:
\textbf{1}) Modality Importance: it measures a model's overall importance of each modality as a whole; and \textbf{2}) 
Modality-Specific Feature Importance (MSFI): it 
measures how well the saliency map can localize the modality-specific important features on each modality. Fig.~\ref{fig:eval_outline} outlines the evaluation methods.

All evaluations are conducted on a brain tumor classification task to classify patients with gliomas into low-grade (LGG) or high-grade gliomas (HGG). 
We use the publicly available BraTS brain MRI dataset~\cite{Menze2015} with an input image $X \in \mathbb{R}^{M \times H \times W \times D}$, and a synthesized brain tumor dataset~\cite{Kim2021} with an input image $X \in \mathbb{R}^{M \times H \times W}$, where $M$ is the number of modalities and $H,W,D$ are width, height, and depth for each single modality MR image. Both datasets include $M=4$ MRI modalities (or pulse sequences) of T1, T1C (contrast enhancement), T2, and FLAIR.



\subsection{Modality Importance (MI)}\label{mod_shapley}
As stated in Section~\ref{intro}, each modality may carry different information for a model's prediction. We use Modality Importance to abstract and simplify such notion. 
It uses importance scores to prioritize how critical a whole modality is to the overall prediction. 
To determine the ground-truth Modality Importance, we use Shapley value from cooperative game theory~\cite{RM-670-PR}, due to its desirable properties such as efficiency, symmetry, linearity, and marginalism. In a set of $M$ modalities, Shapley value treats each modality $m$ as a player in a cooperative game play. It is the unique solution to fairly distribute the total contributions (in our case, the model performance) across each individual modality $m$. 

\paragraph{Shapley value-based ground-truth} We define the modality Shapley value $\varphi_m$ to be the ground truth Modality Importance score for a modality $m$.
It is calculated as: 
\begin{align*}
    \varphi_m(v)=\sum_{c \subseteq \mathcal{M} \backslash\{m\}} \frac{|c| !(M-|c|-1) !}{M !}(v(c \cup\{m\})-v(c))
\end{align*}
where $v$ is the modality-specific performance metric, and $\mathcal{M} \backslash\{m\}$ denotes all modality subsets not including modality $m$. In our evaluation, $v$ is calculated as the prediction model accuracy on the test set. To calculate the performance due to a subset of modalities, we set all values in a modality that is not included in the subset as $0$. We denote such modality Shapley value as $\varphi_m^{\text{mod}}$. We also explored a variant setting where we replace the non-zero values of an ablated modality by sampling from the non-lesion regions. Since such resultant Shapley values had the same ranking and magnitude across modalities, we adopt the setting of zero replacement for simplicity.

\paragraph{Evaluation} To compare the saliency maps' modality importance value with the aforementioned ground-truth, for each saliency map method, we generate and post-process (Section~\ref{postproc}) the saliency maps,
and calculate the \textit{estimated modality importance} as the sum of all positive values of the saliency map for that modality. \textit{MI Correlation} measures the MI ranking agreement between the ground-truth and the \textit{estimated MI}, calculated using Kendall's Tau-b correlation on the test set.



\subsection{Modality-Specific Feature Importance (MSFI)}

The above Modality Importance prioritizes the important modality, but it is a coarse measurement and does not examine the particular image features within each modality. To inspect the fine-grained features within each modality image, we further propose the MSFI (Modality-Specific Feature Importance) metric. It combines two types of ground-truth information:  the modality prioritization information MI, and the feature localization masks/bounding boxes. MSFI is the portion of saliency map values inside the ground truth feature localization mask for each modality, weighted by the normalized MI value $\varphi_m$.

\begin{align}
\hat{\text{MSFI}} &=\sum_{m}  \varphi_m \frac{ \sum_i \mathbbm{1}
( L_m^i >0 ) \odot S_m^i }{ \sum_i 
S_m^i  }\\
\text{MSFI} &= \frac{\hat{\text{MSFI}}} {\sum_{m} \varphi_m}
\end{align}

where $S_m$ is the saliency map for modality $m$, with $i$ denoting the spatial location. $L_m$ is the ground truth feature localization masks/bounding boxes for modality $m$, with $L_m^i>0$ outlining the spatial location of the feature. $\mathbbm{1}$ is the indicator function that selects the saliency map values within the feature mask. 
$ \varphi_m$ is the ground truth Modality Importance value for modality $m$ that is normalized to range $[0,1]$. $\hat{\text{MSFI}}$ is unnormalized, and ${\text{MSFI}}$ is the normalized metric with the range of $[0, 1]$. In our evaluation, we use $S_m^i = \mathbbm{1}(S_m^i >0) \odot \hat{S}_m^i $, where $S_m^i$ is the saliency map that only includes positive values. A higher MSFI score (Fig.~\ref{fig:hm}) indicates a saliency map that better captures the important modalities and their localized features. 
Unlike other common metrics such as IoU, MSFI is less dependent on either saliency map signal intensity, or area of the ground truth localization mask, which makes it a robust metric. Next, we describe our evaluation experiments of applying MSFI on a real dataset (BraTS) and a synthetic dataset.

\subsubsection{MSFI Evaluation on a Medical Image Dataset of Real Patients}
We use the same BraTS data and model as in Section~\ref{mod_shapley}.
To make the ground truth represent the  modality-specific feature localization information, we slightly change the way to compute the modality Shapley value $\varphi_m$. We define $\varphi_m^{\text{feat}}$ to be the importance of the localized feature on each modality. Specifically, instead of ablating the modality as a whole to create a modality subset $c$, we zero-ablate only the localized feature region defined by the segmentation map. We calculate MSFI score with the new ground truth $\varphi_m^{\text{feat}}$.



\subsubsection{MSFI Evaluation on a Synthesized Dataset with Known Ground Truth}\label{syn}

To better control the Modality Importance ground truth, in this second evaluation experiment, we use a synthetic multi-modal medical image dataset on the same brain tumor HGG versus LGG classification task.
We use the GAN-based (generative adversarial network) tumor synthesis model developed by Kim et al.~\cite{Kim2021} to control the shape and coarseness of the tumor textural patterns, and generate two types of tumors mimicking LGG and HGG (Appendix Fig.4) as well as their segmentation maps. The two controllable features are discriminative for tumor grading based on the literature~\cite{Cho2018}. 

To make the discriminative features modality specific, we select two modalities, and assign them different weights to control the degree at which the discriminative features align or agree with the ground-truth labels. The unselected modalities have 0 weights, i.e., they do not contain class discriminative features. For the selected two modalities, we set tumor features on T1C to have 100\% alignment with the ground-truth label, and FLAIR to have a probability of 70\% alignment, i.e., the tumor features on FLAIR corresponds to the correct label with 70\% probability. In this way, the model can choose to pay attention to either the less noisy T1C modality, or to the more noisy FLAIR modality, or to both. To determine their relative importance as the ground truth Modality Importance, 
we test the well-trained model on two datasets:

    \textbf{TIC dataset:} The dataset shows tumors only (without brain background) 
    on all modalities. And the tumor visual features has \textbf{100\%} alignment with ground-truth on \textbf{T1C} modality, and 0\% alignment on \textit{FLAIR}. Its test accuracy is denoted as $\text{Acc}_{\text{T1C}}$.
    
    \textbf{FLAIR dataset:} The dataset shows tumors only  
    on all modalities. And the tumor visual features has \textbf{100\%} alignment with ground-truth on \textbf{FLAIR} modality, and 0\% alignment on \textit{T1C}. Its test accuracy is denoted as $\text{Acc}_{\text{FLAIR}}$.
    
The performance $A_{\text{T1C}}$ and $A_{\text{FLAIR}}$ indicate the degree of model reliance on that modality to make prediction. We then use them as the ground truth Modality Importance.

\section{Physician User Study}\label{user_study}
In addition to the above computational ground-truth, we conduct a user study with doctors to acquire the domain expert ground-truth. The recruited neurosurgeon was introduced to the AI assistant on classifying LGG versus HGG. The doctor was presented two MRI cases and their corresponding saliency maps from the 16 evaluated methods. The doctor was asked to estimate a percentage of match between highlighted saliency areas and his/her clinical judgment on the important features to classify LGG/HGG, and comment on each one. Currently we have recruited one neurosurgeon, and efforts to involve additional neurosurgeons/neuro-radiologists in the study are ongoing. The user study is approved by Research Ethics Board of the university (Study No.: H20-03588). The interview question list is in the Appendix.

\section{Implementation Details}
\subsection{Data and Model}
Our methods are validated on a real clinical dataset --- BraTS 2020 dataset\footnote{Multimodal Brain Tumor Segmentation Challenge \href{http://www.med.upenn.edu/cbica/brats2020/data.html}{http://www.med.upenn.edu/cbica/brats2020/data.html}} and a synthetic dataset for sanity check~(as described in Section~\ref{syn}). The multi-modal images in BraTS dataset are pre-registered and pre-processed. The BraTS dataset includes tumor grade labels of LGG or HGG 
(with the number of cases of LGG: HGG = $76: 293$), and tumor segmentation masks.

Since post-hoc saliency map methods need to explain an already trained model, 
we built a VGG-like 3D CNN that receives multi-modal 3D MR images $X \in \mathbb{R}^{4 \times 240 \times 240 \times 155}$.  We used five-fold cross validation. We denote the model tested on fold $i$ as Fold $i$ model, for $i\in \{1,2,\dots,5\}$. From the remaining data in the other folds, we took around 20\% for validation and the rest for training.
We used weighted sampler to handle the imbalanced data, with a learning rate $= 0.0005$, batch size = 4, and training epoch of 32, 49, 55, 65, 30 for each fold selected by the validation data.
The accuracies of the five folds are $87.81 \pm 3.40\%$~(mean$\pm$std)~(Table~\ref{tab:mod_shapley}). 


For the synthetic brain tumor dataset, we first fitted a classification model the synthesized brain images by fine-tuning the pre-trained DenseNet121 that receives 2D mutli-modal MRI input slices of $X \in \mathbb{R}^{4 \times 256 \times 256}$, with the number of cases of  $\text{LGG}:\text{HGG} = 1:1$.
We used the same training strategies as described above. The model achieves $95.70 \pm 0.06\%$ accuracy on the test set. The saliency maps were generated on a test set with the same ground-truth alignment probability as its training set.  

We used PyTorch\footnote{\href{http://pytorch.org}{http://pytorch.org}} and Monai API\footnote{\href{http://monai.io}{http://monai.io}} for model training, and Captum\footnote{\href{http://captum.ai}{http://captum.ai}} for generating post-hoc saliency maps.  The computational efficiency/speed reported in Fig.~\ref{fig:results} were the inverse normalized log mean computational time using a computer with 1 GTX Quadro 24GB GPU and 8 CPU cores.

\subsection{Post-Processing Saliency Maps} \label{postproc}
Before evaluating and visualizing the saliency maps, we normalized them by first capping the top 1\% outlying values (following~\cite{smilkov2017smoothgrad}). We focused on the positive values of the saliency maps as they are interpreted as the evidence towards the model decision (a.k.a. importance scores), thus set the negative values (features against decision) to zeros. We normalized the values of the saliency maps to $[0,1]$ for easier visualization and comparison. 


\begin{figure*}[!tb]
    \centering
    \includegraphics[width=0.9\textwidth]{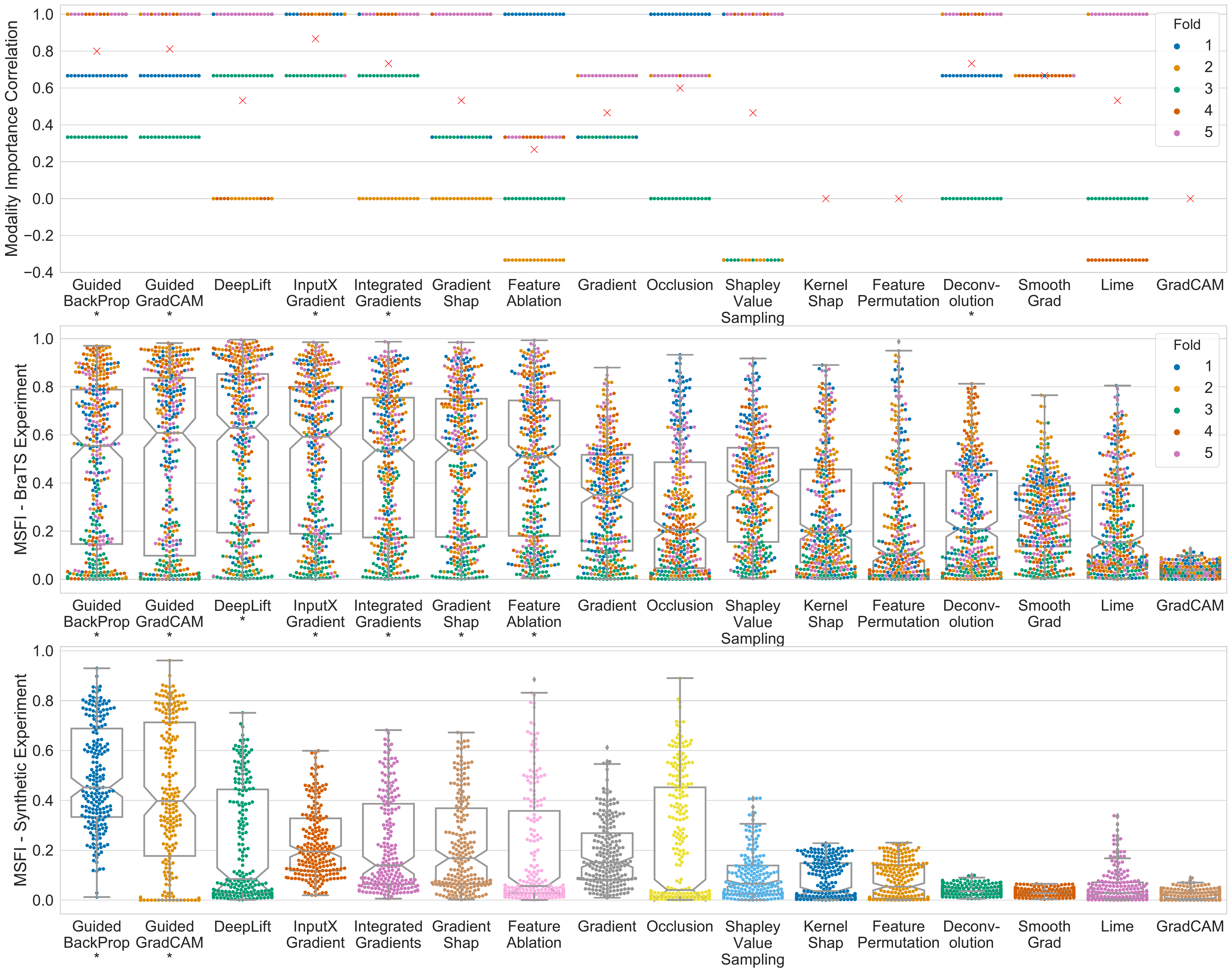}
    \caption{\textbf{Modality Importance correlation and MSFI scores of the evaluated 16 saliency map methods.} 
    The three swarm plots show the evaluation score distribution for MI correlation (top), MSFI on BraTS dataset (middle), and MSFI on synthetic dataset (bottom). X-axis shows each saliency map method. Y-axis shows the evaluation metric, and a higher score means better alignment of a saliency map with the clinical requirements on modality prioritization (MI) and feature localization (MSFI). Each dot is a data sample in the test set, with means indicated by  \textcolor{red}{$\times$} (top), or medians indicated by box plot (middle \& bottom). 
    There was a statistically significant difference among the 16 saliency map methods for each subplot as determined by Friedman test, and the top methods have their name marked with $*$ (determined by not significantly different from the top two means using post-hoc Nemenyi test).}
    \label{fig:metric}
\end{figure*}
\section{Evaluation Result and Discussion}

\subsection{Modality Importance}
The ground-truth modality Shapley value $\varphi_m^{\text{mod}}$ of each modality is shown in Table~\ref{tab:mod_shapley}. We notice Fold 3 model has a different modality importance ranking, and the importance magnitude is significantly smaller than the rest models. This may be due to the model is not well-trained (its performance is the worst). Fig.~\ref{fig:metric}-top shows the ranking similarity between the ground truth and the saliency maps, and the summarized mean is visualized in Fig.~\ref{fig:results}. Except the saliency map methods (GradCAM, KernalSHAP, Feature Permutation) that can only generate one same map for all modalities, the majority saliency map methods could correctly reflect the important modality for the model's decision in general, but with large variations on individual data points.

\begin{table}[!t]
    \centering
    \resizebox{\columnwidth}{!}{%
    \begin{tabular}{l|rrrrr}
    \toprule
      & Fold 1 & Fold 2& Fold 3 & Fold 4 & Fold5    \\
     \midrule
     T1     & 0.03 & -0.04& \textbf{0.028} &-0.07 &-0.01\\
     T1C& \textbf{0.55}  & \textbf{0.53} & \textit{0.012} & \textbf{0.32} &\textbf{0.37}\\
     T2 & -0.04 & 0.05 &-0.007 &0.14 &0.13\\
     FLAIR & \textit{0.16} & \textit{0.16} & -0.005 &\textit{0.24} &\textit{0.21}\\
     \midrule
     Acc & 90.54\% &  90.54\% & 82.43\% & 85.14\% & 90.41\%\\
     \bottomrule
    \end{tabular}
    }
    \caption{\textbf{The Modality Importance ground truth $\varphi_m^{\text{mod}}$ for each model on BraTS dataset}. The first and second important modalities are bolded and italicized respectively. We also list the individual model performance on test set accuracy.}
    \label{tab:mod_shapley}
\end{table}

\subsection{Modality-Specific Feature Importance}
\subsubsection{MSFI Results on BraTS Dataset}
The Spearman correlation between doctor's rating and MSFI score is 0.53 (p=0.001), indicating a moderate correlation between the computational metric and domain expert ground truth. Fig.\ref{fig:hm} gives some qualitative visualization of saliency maps with the two scores\footnote{The 3D saliency maps used in the physician user study can be viewed at: \href{http://vimeo.com/channels/1705604}{http://vimeo.com/channels/1705604}}. A full visualization of all saliency maps and their corresponding doctor ratings/MSFI scores are in the Appendix.

The ground-truth modality Shapley value $\varphi_m^{\text{feat}}$ of each modality is shown in Table~\ref{tab:feat_shapley}. It has the same trend as the $\varphi_m^{\text{mod}}$ value. And the rankings of both $\varphi_m$ align with doctor's prior clinical knowledge, as quoted by the doctor in the user study:
\begin{displayquote}
\textit{``Many of us just look at FLAIR and T1C. 
90\% of my time is on the T1C, and then I will spend 2\% on each of the other modalities.''} 
\end{displayquote}

In addition, doctors use such modality prioritization information and modality-specific feature localization to judge the quality of saliency maps. These clinical requirements on saliency maps are incorporated into the MI and MSFI metrics. 
\begin{displayquote}
\textit{``This one (Feature Ablation saliency map) is not bad on the FLAIR (modality), it (the tumor) is very well detected. I wouldn't give it a perfect mark, because I would like it to prioritize the T1C (modality) instead. But I'll give it (a score of) 75 (out of 100).''} 
\end{displayquote} 

\begin{table}[!t]
    \centering
    \begin{tabular}{l|rrrrr}
    \toprule
     Modality & Fold 1 & Fold 2& Fold 3 & Fold 4 & Fold5    \\
     \midrule
     T1     & 0.01 & -0.01& -0.01 &-0.02 &-0.001\\
     T1C& \textbf{0.22}  & \textbf{0.36} & \textbf{0.09} & \textbf{0.14} &\textbf{0.191}\\
     T2 & -0.02 & 0.04 & 0.01 &0.02 &-0.003\\
     FLAIR & \textit{0.17} & \textit{0.12} & \textit{0.03} &\textit{0.10} &\textit{0.060}\\
     \bottomrule
    \end{tabular}
    \caption{\textbf{The modality-specific feature importance ground truth $\varphi_m^{\text{feat}}$ for each model on BraTS dataset}.}
    \label{tab:feat_shapley}
\end{table}

The average MSFI score for each saliency map method is in the middle to the lower range, with a large variation among individual test data (Fig.~\ref{fig:results}, Fig.~\ref{fig:metric}-middle). Both the computation evaluation and the doctor user study indicate the existing off-the-shelf saliency map methods may not fulfill the clinical requirements on modality-specific feature localization.
Even for methods that have the best mean MSFI scores (around 0.5) such as DeepLift or GuideGradCAM, the MSFI score for individual data points vary across the full range from 0 to 1. Such performance fluctuation of saliency maps may be due to model training (most green dots have very low MSFI score, which is from the least performance model), prediction uncertainty, or weakness of the saliency map method. In our user study, the doctor relied on saliency maps to judge model's prediction quality. Given the large variants, it is unknown whether saliency map can reliably support clinical users' judgment on decision quality~\cite{viviano2021saliency}, and further study needs to inspect the cause of saliency map performance variation.

\subsubsection{MSFI Results on Synthesized Dataset} 
Based on the synthetic dataset setting and the test result $\text{Acc}_{\text{T1C}} = 0.99$, $\text{Acc}_{\text{FLAIR }} = 0$, the ground truth Modality Importance is 1 for T1C, and 0 for the rest modalities. Using this and the segmentation maps, the calculated MSFI score distribution for the test set is shown in Fig.~\ref{fig:metric}-bottom. Most of the saliency map methods can not correctly highlight the ground truth tumor regions on T1C modality, with \textit{GuidedBackProp} and \textit{GuidedGradCAM} outperform the rest.

\section{Conclusion}
Explainable AI is an indispensable component when implementing AI as clinical assistant on medical image-related tasks. In this work, we propose a clinically important but technically ignored problem: explaining model's decisions on multi-modal medical images to clinical users. We propose two evaluation metrices MI (Modality Importance) and MSFI (Modality-Specific Feature Importance). The MI metric indicates the alignment of clinical evidence relying on a certain modality, and further enables the calculation for MSFI that measures saliency map quality with respect to clinical requirements on modality prioritization and feature localization. MSFI metric can be used to evaluate and select the most clinically relevant saliency map method for deployment.
Our evaluation on BraTS and synthetic datasets reveals that existing saliency map methods may not be capable to reflect such modality-specific feature importance information. 
Future work may incorporate MSFI into the objective function, and propose new saliency map methods to precisely reflect the model's learned representations and the clinical prior knowledge.

\section*{Acknowledgement}
We thank Yiqi Yan for his earlier work on Brain Tumor Classification task. We thank Sunho Kim for his generous support to applying his work~\cite{Kim2021} on brain tumor MRI synthesis. We thank Shahab Aslani for the helpful discussions. This study is funded by BC Cancer Foundation-BrainCare BC Fund.
%
%

\nocite{langley00}

\bibliography{xai_eval}
\bibliographystyle{icml2021}





\end{document}


\twocolumn[
\icmltitle{One Map Does Not Fit All: \\
Evaluating Saliency Map Explanation on Multi-Modal Medical Images}




\begin{icmlauthorlist}
\icmlauthor{Weina Jin}{to}
\icmlauthor{Xiaoxiao Li}{goo}
\icmlauthor{Ghassan Hamarneh}{to}
\end{icmlauthorlist}

\icmlaffiliation{to}{School of Computing Science, Simon Fraser University, Burnaby, Canada}
\icmlaffiliation{goo}{Department of Electrical and Computer Engineering, The University of British Columbia, Vancouver, Canada}

\icmlcorrespondingauthor{Weina Jin}{weinaj@sfu.ca}

\icmlkeywords{Interpretable Machine Learning, Medical Imaging Analysis}

\vskip 0.3in
]



\printAffiliationsAndNotice{}  

\section{The Evaluated Saliency Map Methods}\label{xai_method}
We select 16 commonly used activation-, gradient-, and perturbation-based saliency map methods. Except several methods such as CAM, Grad-CAM, feature permutation, and kernel SHAP, the rest can be extended to the multi-modal image input space.
We briefly review the evaluated saliency map methods. 

\subsection{Activation-Based Saliency Map Methods}
\textbf{CAM} (Class Activation Mapping)~\cite{7780688} generates a saliency map for a prediction by aggregating the internal activations of a neural network layer by weighting each neuron's weights in that layer to the final decision layer.

\textbf{Grad-CAM} is similar to CAM but replaces weights with gradients of the target prediction with respect to the activation map. We only include Grad-CAM in our evaluation, because it does not require special model architecture as CAM does.

For activation-based methods including CAM and Grad-CAM, because the activation maps at a deeper layer could not reflect the modality-specific information which is aggregated at the first convolutional layer, the output saliency map is a single-modality image which is not modality-specific. We copy such a saliency map to all modalities to compare it with other methods.

\subsection{Gradient-Based Saliency Map Methods}


\textbf{Gradient} reflects how quickly the output changes when input changes~\cite{simonyan2014deep}.


\textbf{Input $\times$ Gradient} multiples the input to gradient signal to approximate first-order Taylor~\cite{shrikumar2017just}.
Compare with Gradient, it tends to produce sharper saliency maps. Since \cite{shrikumar2017just} showed layer-wise relevance propagation (LRP) is equivalence to Input $\times$ Gradient when all activations are piece-wise linear and biases are included, we only include Input $\times$ Gradient in the evaluation.

\textbf{SmoothGrad} smooths the noisy gradient signals by averaging the saliency maps for an input and its random neighbourhood samples~\cite{smilkov2017smoothgrad}.

\textbf{De-convolution} modifies the gradient computation rule at ReLU activation function. Instead of back-propagating non-negative \textit{input} gradients as in the vanilla Gradient method, De-convolution only back-propagates non-negative \textit{output} gradients~\cite{10.1007/978-3-319-10590-1_53}. 

\textbf{Guided Backpropagation} combines Gradient and De-convolution methods by back-propagating \textit{input} and \textit{output} gradients that are both non-negative~\cite{springenberg2015striving}. 

\textbf{Guided Grad-CAM} computes the element-wise product between Guided Backpropagation and the up-sampled \& broadcasted Grad-CAM signal~\cite{8237336}.

\textbf{Integrated Gradient} approximates the path integral of gradients along the straight line from a neutral baseline input to the target input~\cite{10.5555/3305890.3306024}. 

\textbf{DeepLIFT} explains the prediction difference from a noninformative baseline~\cite{10.5555/3305890.3306006}. DeepLIFT and Integrated Gradients both address the gradient ``saturation'' problem, but DeepLIFT is a faster approximation of Integrated Gradient as it is a modification of the gradient backpropogation rule.

\textbf{Gradient SHAP} approximates Shapley values by computing the expectation of gradients~\cite{NIPS2017_8a20a862}.

\subsection{Perturbation-Based Saliency Map Methods}


\textbf{Occlusion} occludes part of the image with a sliding window, and averaged output differences as the feature attribution~\cite{10.1007/978-3-319-10590-1_53,DBLP:conf/iclr/ZintgrafCAW17}. In our implementation, the occlusion is done modality-wise to generate saliency maps that are modality specific. The occluded regions are replaced by values drawn from a normal distribution with the same mean and standard deviation of the given input modality. We experimented with different sliding window and stride size to balance between saliency map resolution and computational time.

\textbf{Feature Ablation} is similar to Occlusion, but occludes the individual image features rather than using a sliding window. In our implementation, we use modality-wise superpixel segmentation masks as the image features to generate modality-specific saliency maps, and replace the ablated feature with baseline value of $0$s. Cite??

\textbf{Feature Permutation} replaces image feature by shuffling the feature values within a batch, and computes the prediction difference accordingly~\cite{JMLR:v20:18-760}. 

\textbf{LIME} (local interpretable model-agnostic explanation) learns an interpretable model by perturbing and sampling the neighbour data points around the input~\cite{10.1145/2939672.2939778}. 

\textbf{Shapley Value Sampling} A feature's Shapley value is the average marginal feature attribution across all possible feature combination subsets~\cite{RM-670-PR}. Shapley Value Sampling is an efficient sampling method to overcome the expensive enumeration of all possible feature combinations~\cite{CASTRO20091726}.

\textbf{Kernel SHAP} uses the LIME framework to compute Shapley values~\cite{NIPS2017_8a20a862}. Since it only receives one superpixel feature segmentation mask that are shared across modalities, the produced saliency maps are \textit{not} modality-specific.

\section{More Evaluation Results}

\begin{figure*}[h]
    \centering
    \includegraphics[width=0.99\linewidth]{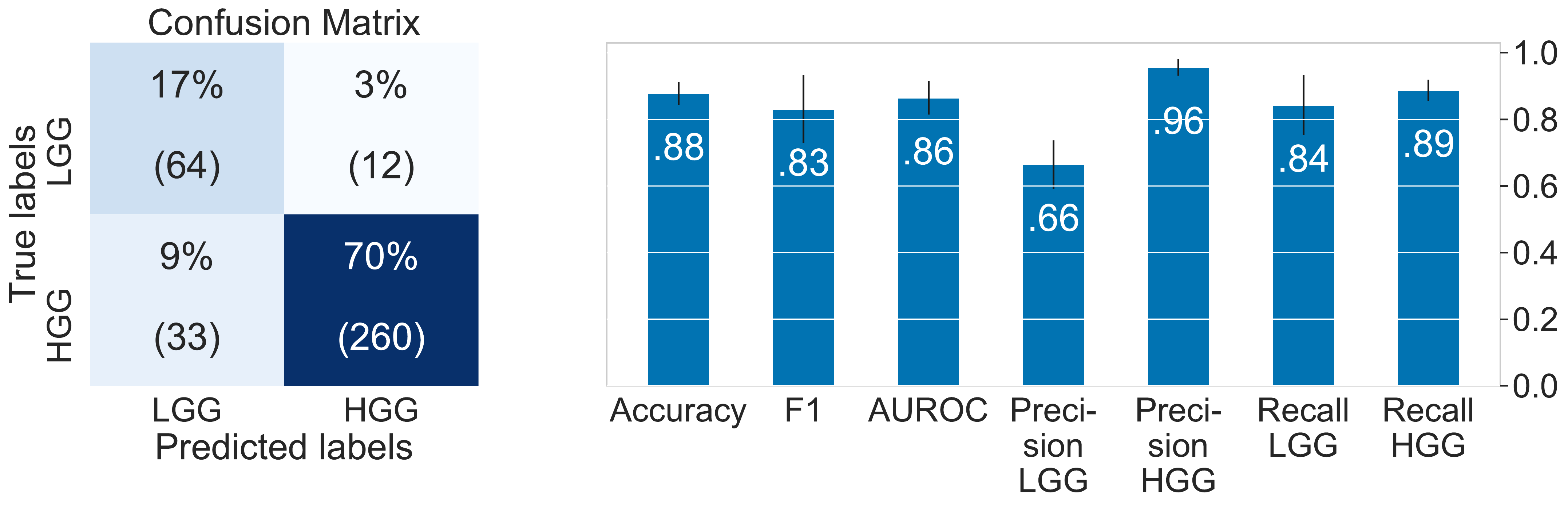}
    \caption{\textbf{Model Performance on brain tumor grade prediction}. Left: Aggregated confusion matrix of testing performance from five-fold models. Right: Mean and standard deviation of the model performance metrics.}
    \label{fig:acc}
\end{figure*}

\begin{figure*}[h]
    \centering
    \includegraphics[width=\textwidth]{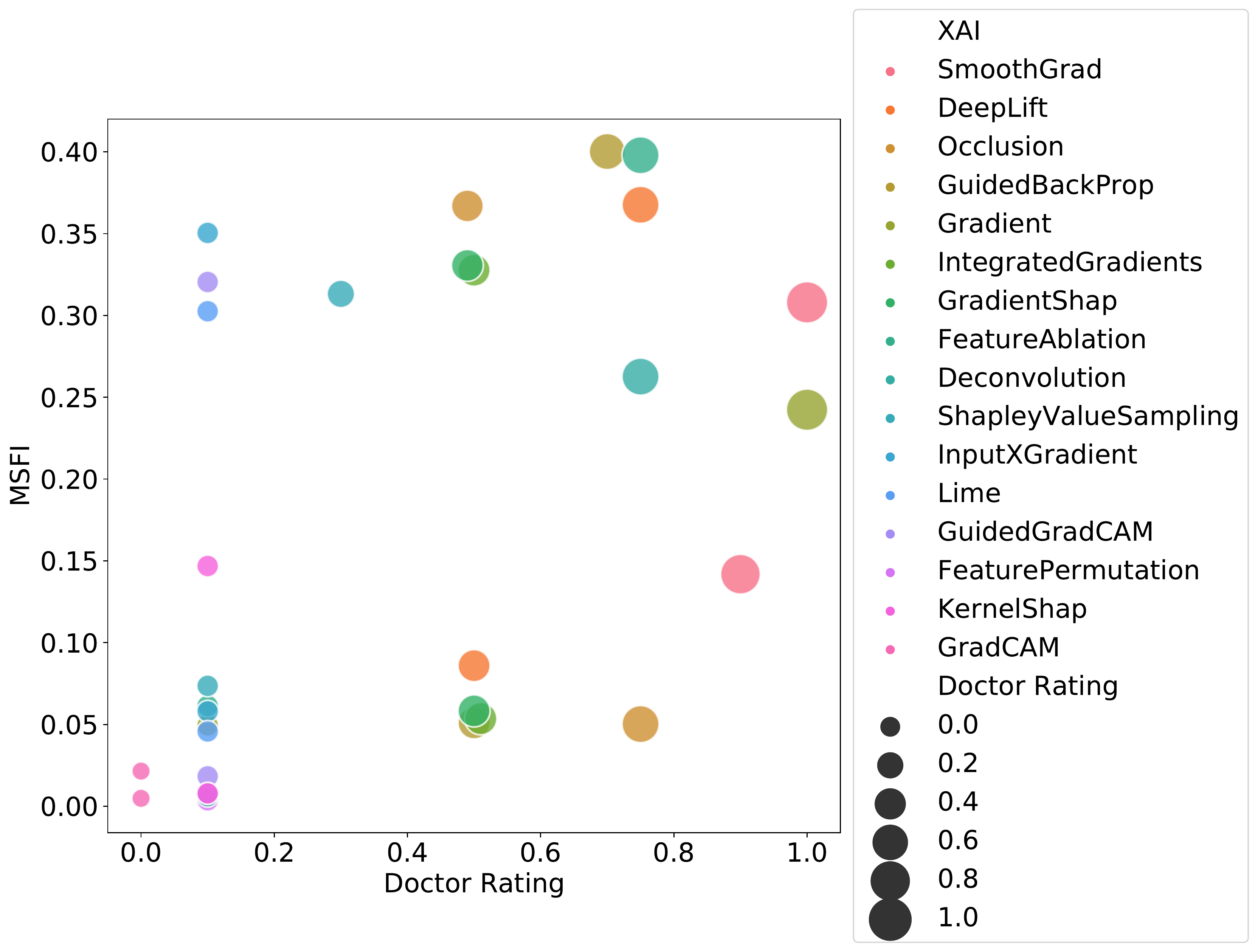}
    \caption{\textbf{Scatter plot showing the correlation of Doctor Rating and MSFI score.} Each method has two doctor ratings on two MRIs. Spearman correlation coefficient = 0.53, p=0.001}
    \label{fig:results}
\end{figure*}


\begin{figure*}%
    \centering
    \subfloat[\centering  HGG]{{\includegraphics[width=\textwidth,height=\textheight,keepaspectratio]{fig/BraTS20_Training_070.pdf} }}%
    \qquad
    \subfloat[\centering LGG]{{\includegraphics[width=\textwidth,height=\textheight,keepaspectratio]{fig/BraTS20_Training_277.pdf} }}%
    \caption{\textbf{The evaluated 16 saliency maps on BraTS dataset/models.} The ground-truth segmentation map is marked as green contour, and the Modality Importance is coded by the intensity of the mask contour. The model predicts correctly on both MRIs.}%
    \label{fig:example}%
\end{figure*}

\begin{figure*}[h]
    \centering
    \includegraphics[width=0.8\linewidth]{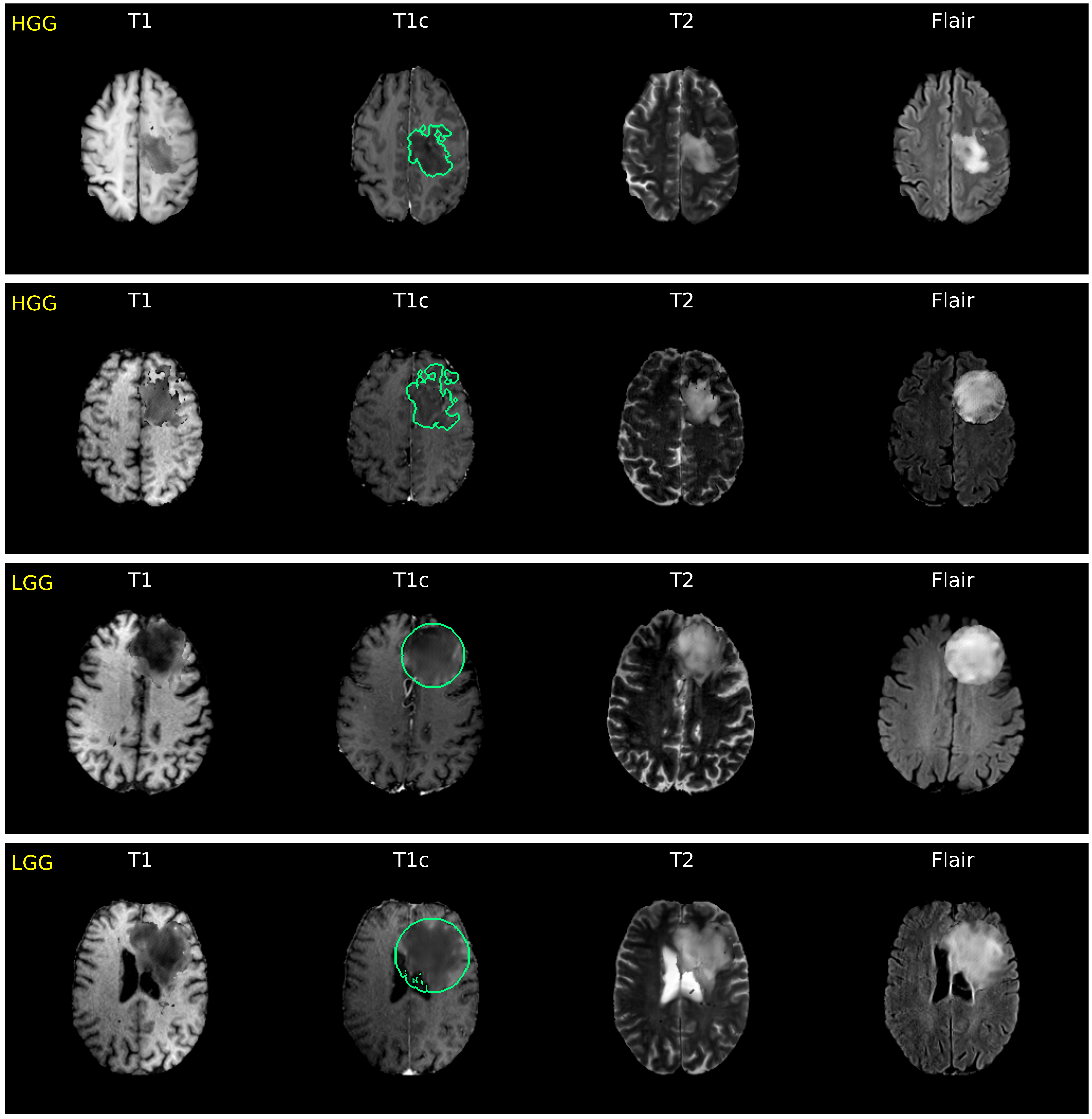}
    \caption{\textbf{The synthesized multi-modal gliomas MRI}.The label of HGG and LGG corresponds to tumor shapes on T1C modality, with the tumor shape outlined in segmentation mask (LGG: round; HGG: irregular). The Flair modality has 70\% alignment with the label, while T1 and T2 modalities are not associated with tumor grade label.}
    \label{fig:hm}
\end{figure*}

\begin{figure*}%
    \centering
    \subfloat[\centering  HGG]{{\includegraphics[width=\textwidth,height=\textheight,keepaspectratio]{fig/58_71.pdf} }}%
    \qquad
    \subfloat[\centering HGG]{{\includegraphics[width=\textwidth,height=\textheight,keepaspectratio]{fig/3_65.pdf} }}%
    \caption{\textbf{The evaluated 16 saliency maps on synthetic dataset/models.}}%
    \label{fig:example}%
\end{figure*}

\begin{figure*}%
    \centering
    \subfloat[\centering  LGG]{{\includegraphics[width=\textwidth,height=\textheight,keepaspectratio]{fig/81_50.pdf} }}%
    \qquad
    \subfloat[\centering LGG]{{\includegraphics[width=\textwidth,height=\textheight,keepaspectratio]{fig/105_46.pdf} }}%
    \caption{cont. \textbf{The evaluated 16 saliency maps on synthetic dataset/models.}}%
    \label{fig:example}%
\end{figure*}

\begin{figure*}[!hb]
    \centering
    \includegraphics[width=\textwidth]{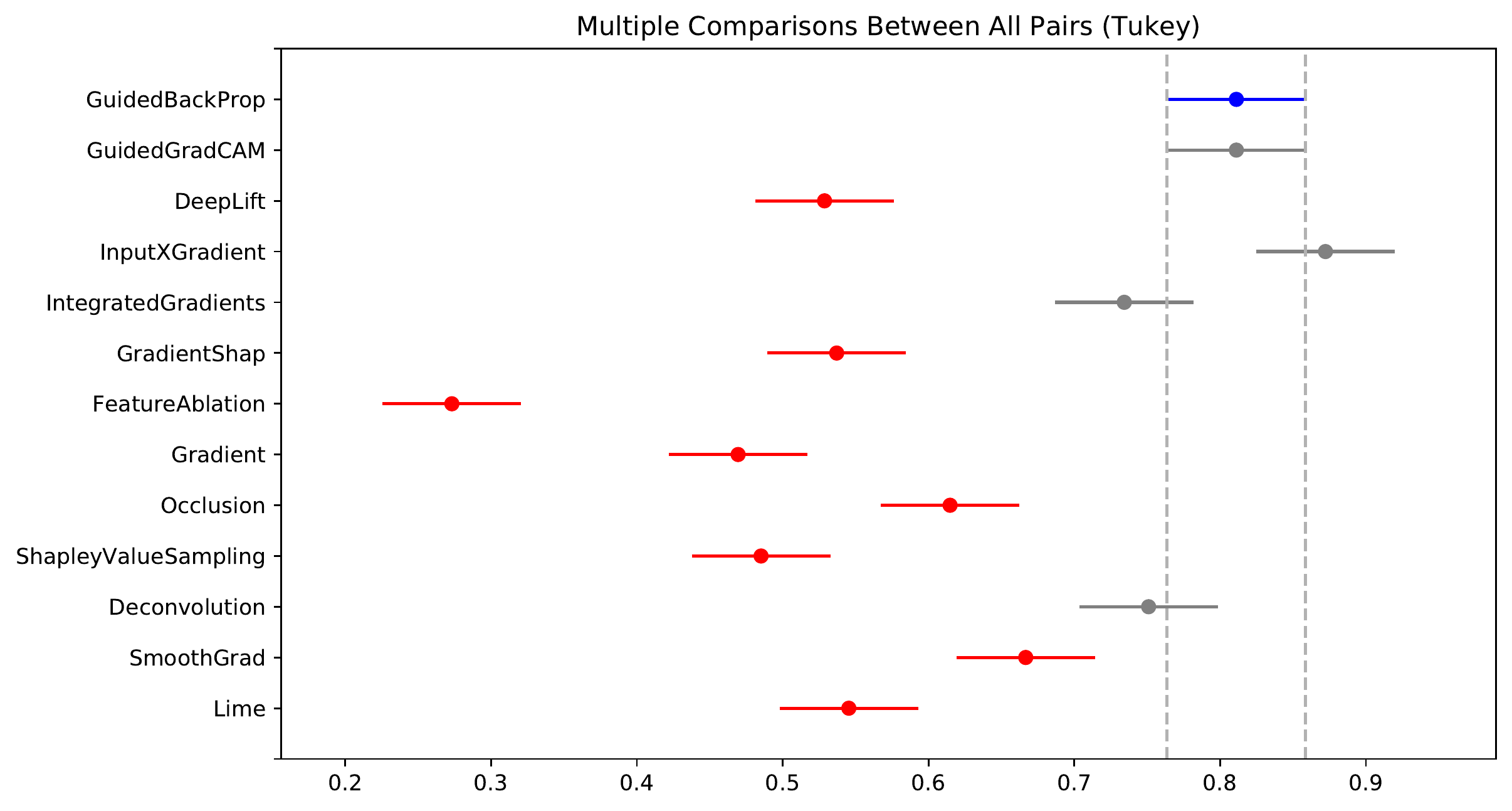}
    \caption{\textbf{Modality Importance Ranking Correlation.} The plot shows means and confidence intervals of Modality Importance Ranking Correlation. The confidence intervals are computed using Tukey's HSD test. The saliency map methods which means are significant different can be determined if their intervals do not overlap.}
    \label{fig:results}
\end{figure*}

\begin{figure*}[!hb]
    \centering
    \includegraphics[width=\textwidth]{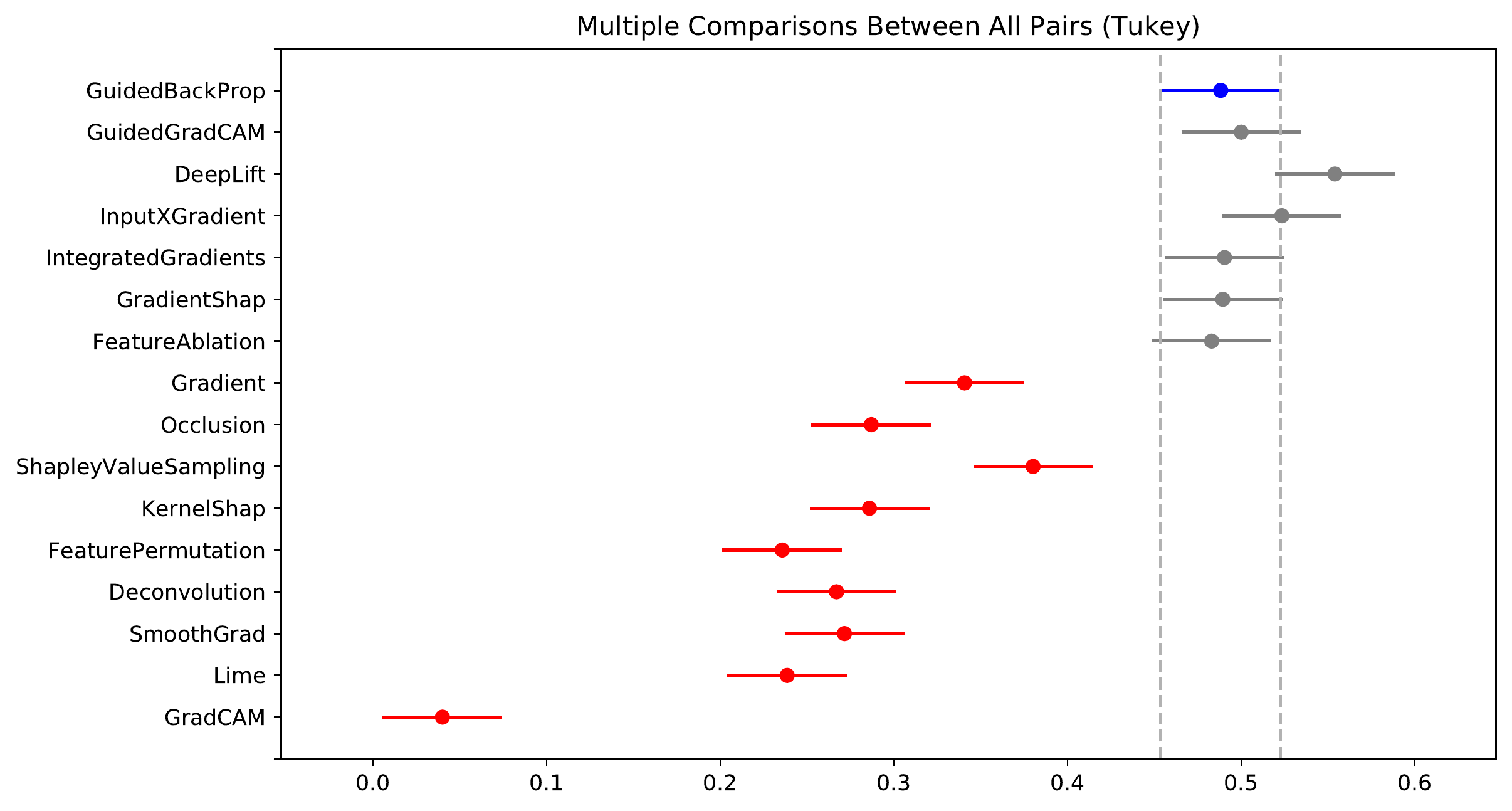}
    \caption{\textbf{MSFI: BraTS Experiment}}
    \label{fig:results}
\end{figure*}

\begin{figure*}[!hb]
    \centering
    \includegraphics[width=\textwidth]{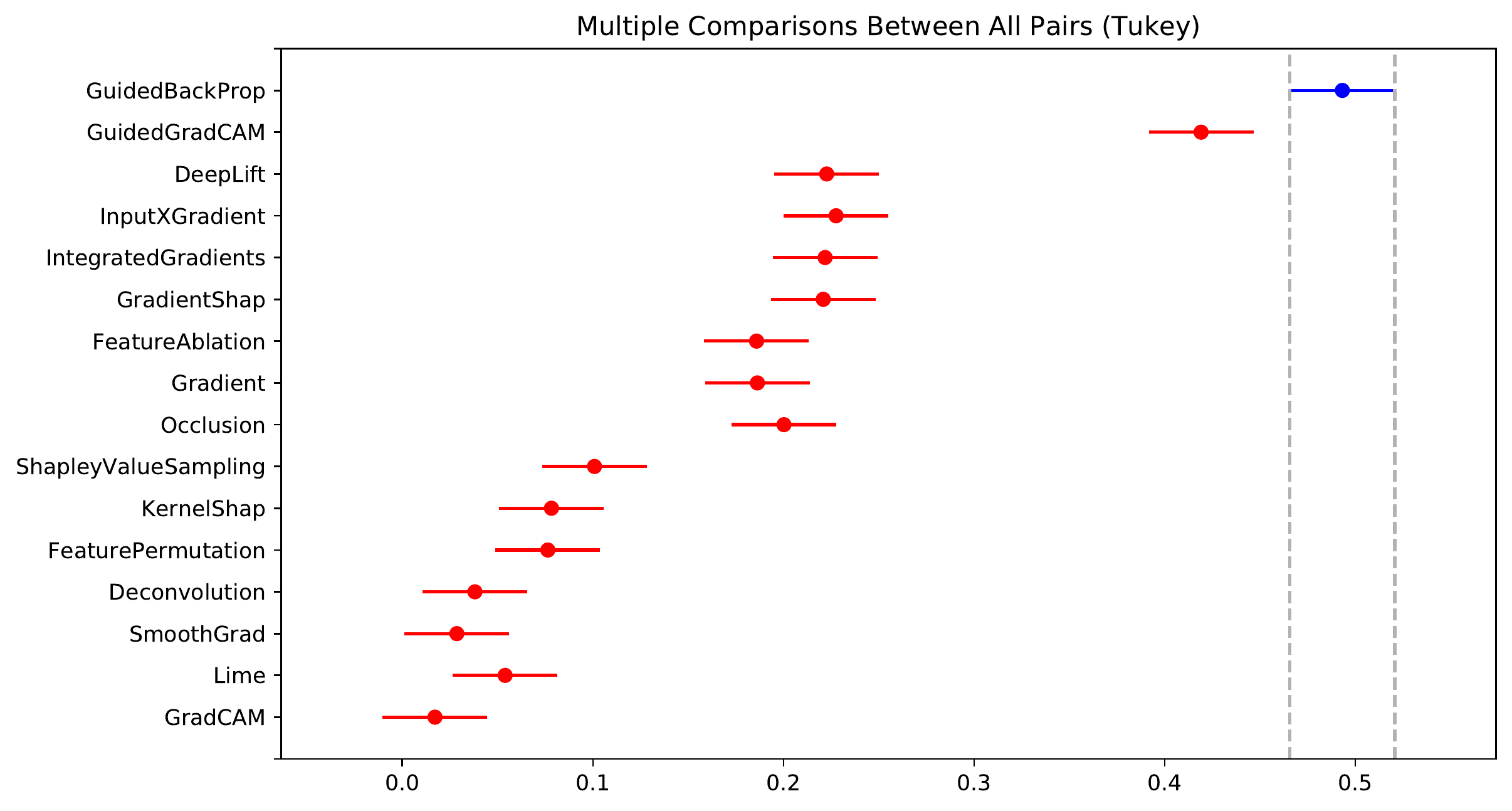}
    \caption{\textbf{MSFI: Synthetic Experiment}}
    \label{fig:results}
\end{figure*}
%
%

\clearpage

\bibliography{xai_eval}
\bibliographystyle{icml2021}



